\documentclass[preprint,12pt]{elsarticle}

\usepackage{amssymb}
\usepackage{amsmath}
\usepackage{amsfonts}
\usepackage{algorithmic}
\usepackage{graphicx}
\usepackage{multirow}
\usepackage{float}
\usepackage{subcaption}
\usepackage{textcomp}
\usepackage{xcolor}
\usepackage{array}
\usepackage{tabularx}
\usepackage{comment}

\newtheorem{remark}{Remark}

\journal{Engineering Applications of Artificial Intelligence}

\begin{document}

\begin{frontmatter}



\title{Data-Driven vs Traditional Approaches to Power Transformer’s Top-Oil Temperature Estimation}



\author[1]{Francis Tembo}
\affiliation[1]{organization={Division of Decision and Control Systems, Department of Intelligent Systems, KTH Royal Institute of Technology}, 
  city={Stockholm},
  country={Sweden}}

\author[2]{Federica Bragone}
\affiliation[2]{organization={Division of Computational Science and Technology, Department of Computer Science, KTH Royal Institute of Technology }, 
  city={Stockholm},
  country={Sweden}}

\author[3]{Tor Laneryd}
\affiliation[3]{organization={Hitachi Energy Research}, 
  city={Västerås},
  country={Sweden}}

\author[1]{Matthieu Barreau}

\author[1]{Kateryna Morozovska}


\begin{abstract}
Power transformers are subjected to electrical currents and temperature fluctuations that, if not properly controlled, can lead to major deterioration of their insulation system. Therefore, monitoring the temperature of a power transformer is fundamental to ensure a long-term operational life. Models presented in the IEC 60076-7 and IEEE standards, for example, monitor the temperature by calculating the top-oil and the hot-spot temperatures. However, these models are not very accurate and rely on the power transformers' properties. This paper focuses on finding an alternative method to predict the top-oil temperatures given previous measurements. Given the large quantities of data available, machine learning methods for time series forecasting are analyzed and compared to the real measurements and the corresponding prediction of the IEC standard. The methods tested are Artificial Neural Networks (ANNs), Time-series Dense Encoder (TiDE), and Temporal Convolutional Networks (TCN) using different combinations of historical measurements. Each of these methods outperformed the IEC 60076-7 model and they are extended to estimate the temperature rise over ambient. To enhance prediction reliability, we explore the application of quantile regression to construct prediction intervals for the expected top-oil temperature ranges. The best-performing model successfully estimates conditional quantiles that provide sufficient coverage.
\end{abstract}




\end{frontmatter}




\section{Introduction}
Power transformers experience loss of life when the efficiency of their insulation system degrades. During normal operation, heat is generated in the copper windings due to their resistance to the flow of electrical current. However in recent years, capacity extension of power grid components such as power lines \cite{MICHIORRI20151713,estanqueiro2018dlr}, cables \cite{michiorri} and power transformers \cite{hamza1, morozovska2021dynamic} is becoming increasingly popular especially for integration of renewable energy sources to the grid \cite{morhel, VIAFORA2019194}. Dynamic capacity extension often leads to increased average temperatures and longer hours of high-temperature operation \cite{bracale2019probabilistic}. Continuous operation at high temperatures typically leads to the physical and chemical deterioration of oil-paper insulation. The degradation of the insulation properties is directly related to the power transformer's remaining lifetime. Since it is quite critical that power transformers can be reliably loaded for safe transmission and distribution of electrical power, monitoring their thermal conditions has been an active area of research over the last decades. 

The IEEE and IEC loading guides \cite{ref1,ref2} have developed several models for estimating a power transformer's remaining lifetime based on top-oil and winding hot-spot temperatures. These models are often used for quickly assessing a power transformer's internal temperatures. However, they have low accuracy and rely on the power transformer's properties for their calculations. This highlights the need for more accurate models that do not depend on power transformer characteristics. With the increasing popularity of machine learning models, there has been a growing interest in research using them for power transformer condition monitoring. Earlier research demonstrated the promising potential of using neural networks for estimating the top-oil temperature for transformers \cite{ref3}. The growing understanding of neural network architectures and increasing computational power have led to sustained interest in investigating the thermal modeling of power transformers.

A recent study investigated nonlinear autoregressive neural networks and support vector machines to predict winding temperature under various loading configurations \cite{ref4}. Autoregressive refers to the nature in which the estimations are made by using the model's previous estimations as input to generate subsequent predictions. The measurements were collected from 9 different power transformers, and the models were shown to have estimations much closer to the measurements compared to those produced by the standard IEEE and IEC models. Additionally, the study investigates the use of wind speed and solar radiation information and how, in some cases, it can improve performance. In addition to predicting the top-oil and winding temperatures, neural networks, particularly physics-informed neural networks (PINNs), can also provide additional information regarding the internal thermal behavior of power transformers  \cite{ref5}. Further study on the heat distribution is presented in \cite{fede1} by investigating the parameters of the equation used and in \cite{fede2} by analyzing the results of PINNs with several grids for the numerical method solution. The degradation of the insulation system applying PINNs for estimating the corresponding critical parameters is studied in \cite{fede3}. In \cite{hamza2} the authors present thermoelectric equations to model power transformers and validate them with real measurements. The same models are tested in \cite{hamza1} for the Anholt offshore wind farm as a use case to give an insight on how to dimension power transformers for offshore wind.



This study investigates time-series forecasting models for estimating the top-oil temperature of an operating power transformer based on historical measurements. Additionally, we explore how quantile regression can be used to determine a range of possible values for the estimate within a prediction interval.

 \section{Methodology}
 \label{sec:methodology}
In this study, we investigate artificial neural networks (ANNs) and a selection of time-series forecasting models available in the open-source Darts Python library \cite{ref6}. The package provides a user-friendly way to explore time-series forecasting with various models. Darts includes a wrapper over popular statistical and neural network models, which can be accessed through a unified API. The package supports univariate and multivariate time-series forecasting, including several ways to handle the external data that is relevant to forecasts. This external information is referred to as covariates, which can either be known from the past as measurements (past covariates) or into the future as forecasts (future covariates). 

\subsection{Standard and Quantile Regression}
Standard regression analysis typically aims to model the relationship between one or more dependent variables $\mathbf{X}$ and a dependent variable $\mathbf{Y}$. The conditional distribution function of $\mathbf{Y}$ given $\mathbf{X}$ can be defined as: 
\begin{equation}
F(y \mid \mathbf{X}=\mathbf{x}) = \mathbb{P}(\mathbf{Y}\leq y \mid \mathbf{X}=\mathbf{x}),
\end{equation}
where $\mathbf{X}=(x_0,x_1,...,x_p)$ represents the set of independent variables. Traditional least-of-squares regression aims to estimate the conditional mean function $\hat{\mathbf{Y}}=E[\mathbf{Y}|\mathbf{X}]$ of a target variable by minimising the squared error: 
\begin{equation}
    L(y, \hat{y}) = (y - \hat{y})^2,
\end{equation}
where $\hat{y}$ is the estimate and $y$ is the target. 

\begin{figure}[H]
\centering
\includegraphics[width=0.9\textwidth]{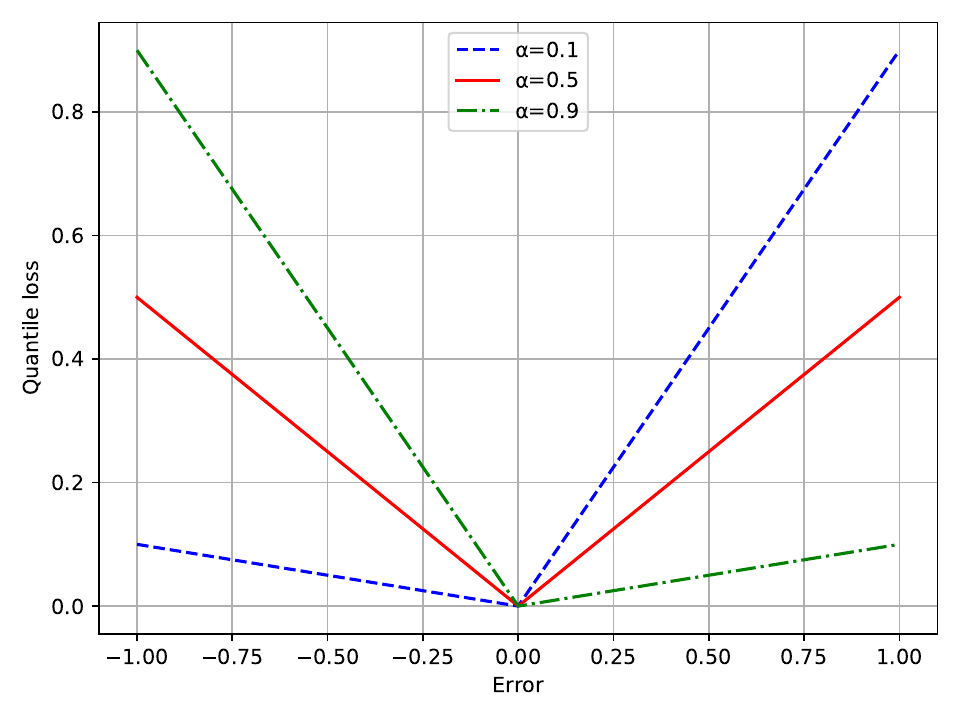}
\caption{Quantile loss function for varying values of $\alpha$.}
\label{quantile_loss_depiction}
\end{figure}

Darts supports probabilistic forecasting, of particular interest to this study, an extension of regression referred to as quantile regression \cite{ref7}. It extends the standard regression by allowing the estimation of any conditional quantile of the target variable. This is achieved by introducing an additional parameter \(\alpha \in [0, 1]\) representing the desired quantile. Where the conditional function for the $\alpha^{\text{th}}$ quantile can be defined by:
\begin{equation}
    Q_{\alpha}(\mathbf{x}) := \inf{\{y \in \mathbb{R} : F(y\mid \mathbf{X}=\mathbf{x}) \geq \alpha \}}
\end{equation}
The quantile loss can then be defined as:
\begin{equation}
L_{\alpha}(y, \hat{y}) =
\begin{cases}
\alpha (y - \hat{y}) & \text{if } y - \hat{y} \geq 0 \\
(\alpha - 1) (y - \hat{y}) & \text{if } y - \hat{y} < 0
\end{cases}
\end{equation}
This can also be compactly written as:
\begin{equation}
L_{\alpha}(y, \hat{y}) = \max(\alpha(y - \hat{y}), (\alpha - 1)(y - \hat{y}))
\end{equation}
Interestingly, similar to how median regression estimates the conditional median by minimizing the symmetrically weighted sum of absolute errors, quantile regression estimates a conditional quantile $\alpha$ of the target variable by minimizing an asymmetrically weighted sum of absolute errors. As depicted by Figure \ref{quantile_loss_depiction}, the weighted sum of absolute errors becomes symmetrical at $\alpha$ = 0.5, and the quantile loss estimates the conditional median:
\begin{equation}
\begin{array}{rl}
L_{0.5}(y, \hat{y})\!\!\!\!\!&= 
\frac{1}{2} \left( \max(y - \hat{y}, 0) + \max(\hat{y} - y, 0) \right) \\
&= \frac{1}{2} |y - \hat{y}|.
\end{array}
\end{equation}
Conversely, for $\alpha$ $(0.5,1]$, the loss is asymmetrical estimations lower than the target are penalized more than estimations higher than the target. The reverse is true for overestimations when $\alpha$ is less than 0.5.

To evaluate the performance of a quantile regression model over a dataset, we use the mean quantile loss (MQL):
\begin{equation}
\text{MQL}_{\alpha} = \frac{1}{n_{\text{s}}} \sum_{i=1}^{n_{\text{s}}} \left(\alpha |y_i - \hat{y}_i|_+ + (1 - \alpha) |\hat{y}_i - y_i|_+ \right)
\end{equation}
where $| x |_+ = \max(x, 0)$ represents the positive part and $n_s$ is the number of samples. Since we are interested in obtaining multiple quantiles, we average the mean quantile loss for all the quantiles. 

Quantile pairs can be used to construct conditional prediction intervals \cite{romano2019conformalized}, \cite{alcantara2023deep}. A conditional prediction interval defines the likelihood that the dependent variable lies within the upper and lower bounds of conditional quantiles. Given the conditional quantile function pairs  $Q_{\alpha_{\text{upper}}}(\mathbf{x})$ and $Q_{\alpha_{\text{lower}}}(\mathbf{x})$ representing the upper  $\alpha_{\text{upper}} = 1 - \alpha/2$ and  lower $\alpha_{\text{lower}} = \alpha/2$ quantiles respectively. The conditional prediction interval can be given as: 
\begin{equation}
    PI(\mathbf{x}) = [Q_{\alpha_{\text{lower}}}(\mathbf{x}), Q_{\alpha_{\text{upper}}}(\mathbf{x})].
\end{equation}
The conditional prediction interval satisfies the following: 
\begin{equation}
\mathbb{P} \left( Y \in PI(\mathbf{X}) \mid \mathbf{X} = \mathbf{x} \right) \geq 1 - \alpha.
\label{prediction_interval_condition_satisfaction}
\end{equation}

Quantile regression and prediction intervals offer a useful extension to typical regression where point estimations are inadequate and a mechanism that offers certainty of predictions is required. The construction of prediction intervals is intelligible and quantiles give additional information about the relation between the dependant and independent variables.

 \subsection{Models considered}

 \subsubsection{Artificial Neural Networks (ANNs)}
 ANNs consist of several layers where the data passes through and is processed \cite{bishop2006pattern}. Each layer comprises multiple neurons connected between the layers, creating fully connected layers. Information is loaded through an input layer, passes through the hidden layers where non-linear transformations are applied, and gets out, after being transformed, through an output layer. Each neuron has a weight $w$ and a bias $b$ assigned to it, and a non-linear transformation is assigned to the weighted sum of the inputs plus the bias as defined by \eqref{eqbias}.
 \begin{equation}
     y = \phi\bigg(\sum_{j=1}^{n} w_jx_j + b\bigg)
     \label{eqbias}
 \end{equation}
 where $y$ is the output, $x_j$ are the inputs for $j=1,...,n$, and $\phi:\mathbb{R}\to \mathbb{R}$ is a non-linear map, defined as the activation function. To train an ANN we define a loss function that calculates the error between the estimated output and the target, which we minimize by an iterative method. The iterative methods use optimization algorithms to update the model's parameters (weights and biases). First, the weights are initialized, and the input is forward propagated through the network to obtain the output. The error is calculated using a loss function and is backpropagated through the network to compute the gradient of the loss function with respect to each weight. The weights are then updated using an optimization algorithm, like Stochastic Gradient Descent (SGD). These steps are repeated until the model converges. 
\subsubsection{Time-series Dense Encoder (TiDE)}
This architecture was introduced as an alternative to transformer-based neural network architectures for long-term time-series forecasting. Instead of implementing the encoder-decoder units with attention, this architecture uses simple, densely connected multilayer perceptrons \cite{ref12}. 

\begin{figure}[H]
\centering
\includegraphics[width=0.95\textwidth]{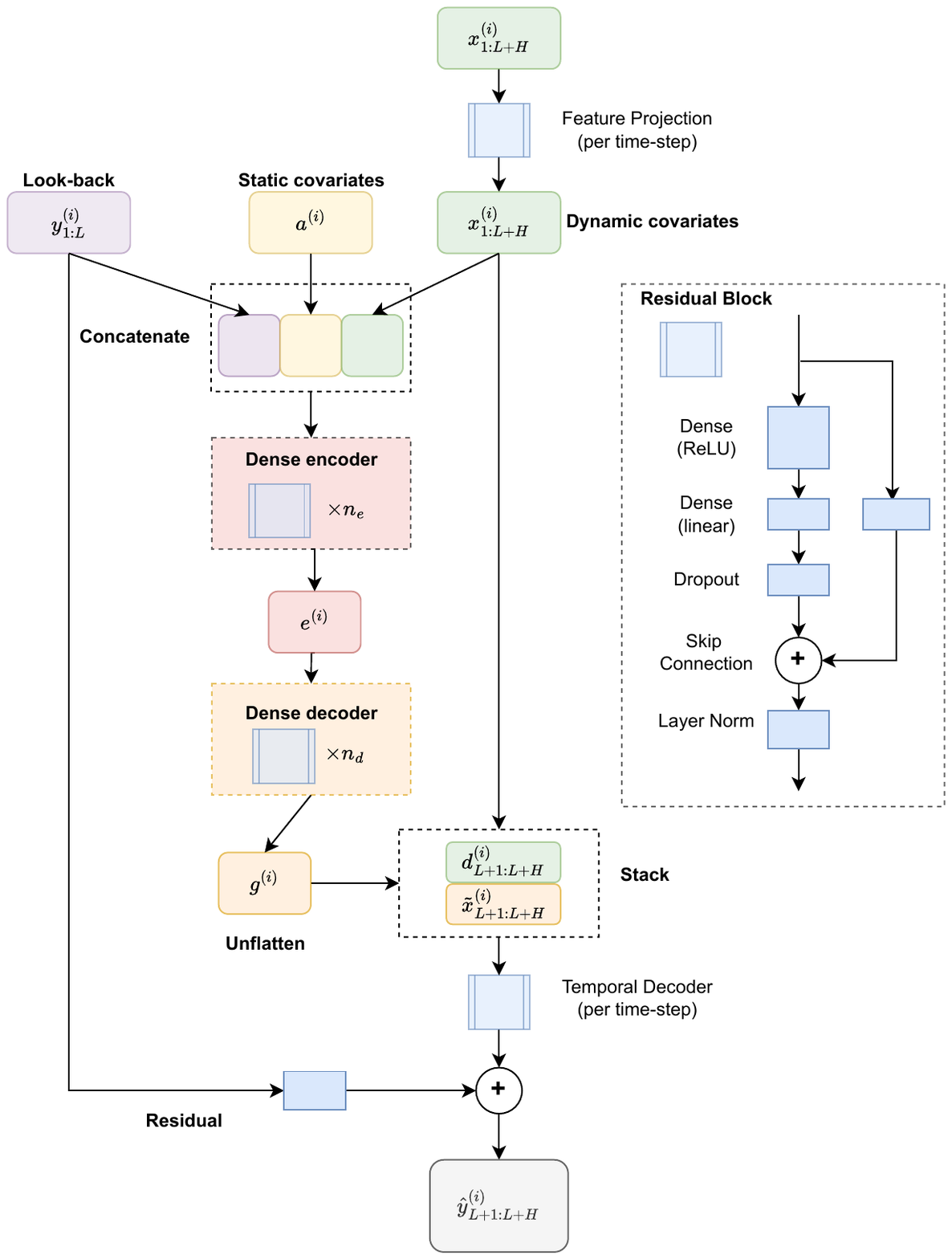}
\caption{Overview of TiDE architecture.}
\label{tide_architecture}
\end{figure}
This approach requires less training time and is faster at inference without sacrificing performance. The encoder’s responsibility is to transform the target series and the covariates into a dense feature representation. TiDE follows the idea that given a dataset with $N$ time-series, forecasting involves mapping the time-series history with relevant information to an estimation of the future:
\begin{equation*}
    \begin{array}{ccl}
        \displaystyle \{ {\mathbf{y}}^{(i)}_{1:L}, {\mathbf{x}}^{(i)}_{1:L+H}, {\mathbf{a}}^{(i)} \}_{i=1}^N 
        & \mapsto & 
        \{ {\hat{\mathbf{y}}}^{(i)}_{L+1:L+H} \}_{i=1}^N, \\
    \end{array}
\end{equation*}
where $\mathbf{y}^{(i)}_{1:L}$ is the $i^\text{th}$ time-series of look-back $L$, $\mathbf{x}^{(i)}_{1:L+H}$ are the dynamic covariates spanning the look-back $L$ and the forecast horizon $H$, and $\mathbf{x}^{(i)}_{1:L+H}$ are the static covariates if available. 

\begin{remark}
    It is important to note that for our application we first explore the estimation of both the top-oil temperature and temperature rise over ambient. In this case, $N=2$ and the dynamic covariates are the ambient temperature and the load factor. When exploring quantile regression, we only estimate the top-oil temperature and $N=1$.
\end{remark}

The first step in the encoding process is a dimensionality reduction operation on the dynamic covariates using a process referred to as feature projection. Given the $r$-dimensional dynamic covariates $ \mathbf x_t^{(i)} \in \mathbb{R}^r$, a residual bloc is defined as
\begin{equation}
    \tilde{ \mathbf {x}}_t^{(i)} = \text{ResidualBlock}(\mathbf {x_t}^{(i)}).
    \label{eqresblock}
\end{equation}
It is used at each time step to map the $\mathbf x_t^{(i)}$ to a lower-dimensional representation of the size $\tilde{r} \ll r$ that represents the temporal width.

In the second step \eqref{eqencoder}, the covariate projections are combined with the target series and its static covariates before it is mapped to an embedding using a dense encoder composed of multiple residual blocks. 
\begin{equation} 
    \mathbf{e}^{(i)} = \text{Encoder}(\mathbf{y}_{1:L}^{(i)}; \tilde{\mathbf{x}}_{1:L+H}^{(i)};\mathbf{a}^{(i)}), 
    \label{eqencoder}
\end{equation}
where $ \mathbf y_{1:L}^{(i)}$ represents the look-back of the $i$-th time-series and $\mathbf a^{(i)}$ the static properties of the time-series often referred to as static covariates. The motivation behind the encoder is to learn the distilled representations that best convey the useful patterns from the input data. The temporal width determines time steps that the model can process at once, i.e. the width of the layers in the covariate projection. The embedding is then passed to the decoder, which generates the forecast estimates for the target time series.

Similarly to encoding, decoding is carried out in two steps as shown in \eqref{eqdecoder}. First, a dense decoder takes the input encoding $ \mathbf e^{(i)}$ and transforms it into a vector $ \mathbf g^{(i)}$ with dimensions $H \times p$ where $p$ is the output dimension of the decoder and $H$ is the horizon length. The vector $\mathbf g^{(i)}$ is later reshaped as a matrix $\mathbf D^{(i)} \in \mathbb{R}^{d \times H}$. The column \(t\), denoted as \(\mathbf d(i)_t\), can be interpreted as the decoded vector for the period of time \(t\) within the horizon, where \(t \in [H]\). 
\begin{equation}
    \begin{array}{l}
        \mathbf g^{(i)} = \text{Decoder}(\mathbf e^{(i)}) \in \mathbb{R}^{p\times H}, \\
        \mathbf D^{(i)} = \text{Reshape}(\mathbf g^{(i)}) \in \mathbb{R}^{p \times H}.
    \end{array}
    \label{eqdecoder}
\end{equation}

In the next step, the temporal encoder unit takes in the estimated forecast from the previous encoder and adopts information related to the projected covariates. For the $t$-th horizon time-step and projected covariates $\tilde{\mathbf x}_{L+t}^{(i)}$:
\begin{equation}
\hat{\mathbf y}_{L+t}^{(i)} = \text{TemporalDecoder}\left(\mathbf d_t^{(i)}; \tilde{\mathbf x}_{L+t}^{(i)}\right) \quad \forall t \in [H].
\end{equation}

In the final step, a global residual connection is used to linearly transform the look-back $\hat{ \mathbf y}_{1:L}^{(i)}$ into a vector of the same size as the horizon. The transformed vector is then added to the prediction $\hat{\mathbf y}_{L+1: L+H}^{(i)}$. 

The novelty of using residual connections and the temporal decoder was shown to give TiDE a performance advantage compared to its transformer-based models on several time-series benchmark datasets. Since the temporal encoder adds information regarding future covariates, this ensures that future changes are considered when making forecast estimates. Furthermore, the model inherits the advantages of incorporating residual connections in a deep neural network. 

\subsubsection{Temporal Convolutional Networks (TCN)}
The TCN is an adaptation of the convolutional neural network to handle sequential data, specifically using dilated convolutions to model temporal dependencies \cite{ref13}. The dilated convolutions allow for large receptive fields without the drawback of an increase in computation complexity. The TCN design philosophy follows two assumptions: 
\begin{itemize}
    \item{Causality as no information is leaked from the future to the past.}
    \item{Arbitrary length sequence-to-sequence mapping through 1D fully convolutional layers.}
\end{itemize}
Causality is ensured through casual convolutions. It is a type of convolution where the operations respect the temporal order of the data. The output at a specific time step $t$ is computed using only the input data available up to and including that time step $t$, avoiding any future information leakage. The arbitrary sequence-to-sequence mapping is achieved by having the next layer the same length as the previous layer. 

The TCN has been shown to have better performance compared to Recurrent Neural Networks (RNNs) and Long Short-Term Memory (LSTM) networks on several tasks \cite{ref13}. Furthermore, the TCN requires less memory as filters are shared across a layer, and parallelism because convolution operations are done in parallel. Similar to the TiDE architecture, TCN employs residual blocks, which allow its layers to learn adjustments related to the identity mapping instead of the complete transformation.


\subsubsection{IEC 60076-7}
The conventional method to calculate the power transformer temperature during operation is outlined in \cite{nordman2003temperature}. The top-oil temperature change is described, according to the IEC 60076-7 standard \cite{ref2}, using the following differential equation:
\begin{equation}
    \bigg[\frac{1+K(t)^2\Psi}{1+\Psi}\bigg]^{\chi}(\Delta T_{or}) = k_{11}\tau_o\frac{dT_o}{dt}+[T_o(t) - T_a(t)].
    \label{eq:IEC}
\end{equation}
where $K(t)$ is the load factor [p.u.]; $\Psi$ is the ratio of the load losses at rated current to no-load losses at rated voltage; $\Delta T_{or}$ is the top-oil temperature rise in steady-state at rated losses [K]; $\chi$ is the exponential power of total losses vs. the top-oil temperature rise; $k_{11}$ is a thermal model constant; $\tau_o$ is the oil time constant [min]; $T_a(t)$ and $T_o(t)$ are the ambient temperature and the top-oil temperature [$^{\circ}$C], respectively. The IEC 60076-7 standard provides an intelligible and formalised approach to determining top-oil temperature. However, the thermal model depends on the specific thermal properties of the power transformer. 

\section{Case Study}
\begin{figure}[ht]
\centering
\includegraphics[width=0.9\textwidth]{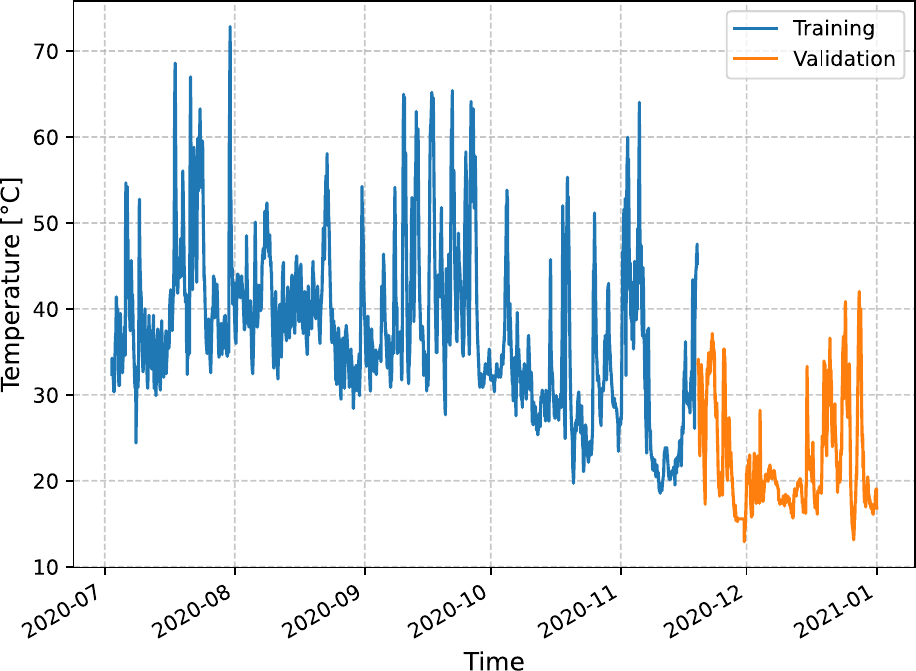}
\caption{Top-oil temperature measurements over the training and validation datasets.}
\label{top_oil_temperature_training_and_validation}
\end{figure}

The data used in this study are measurements taken from an operating power transformer. The measurements are electrical current and top-oil temperature values that are sampled at five-minute intervals for 190 days between 25\textsuperscript{th} June 2020 and 1\textsuperscript{st} January 2021. 
Missing top-oil temperature values due to measurement errors are replaced by the average of adjacent values. Instead of using raw current values, the load factor $K$ is used. The ambient temperature values are acquired from publicly available weather data from the nearest weather station \cite{smhi}. The ambient temperature data is sampled once at the start of every hour. To match the sampling rate of the transformer measurements, ambient temperatures are linearly interpolated between consecutive hourly readings. The ambient temperature data had missing values; instead of further interpolating or completely discarding these values to maintain temporal consistency we retain the remaining data as the validation set. It is important to note that the validation set has samples of lower top-oil temperatures that are not present in the validation set, this is depicted in Figure \ref{top_oil_temperature_training_and_validation}. Table \ref{table:data_splits} gives a detailed description of the splits, roughly only six days and 18 hours worth of data are lost instead of completely disregarding the remaining data. 

\begin{table}[ht]
\renewcommand{\arraystretch}{1.3}
\caption{Data Splits With Corresponding Start and End Dates, Number of Data Points, and Duration}
\label{table:data_splits}
\centering
\begin{tabular}{|l|l|l|p{1.5cm}|p{3cm}|}
\hline
\footnotesize \textbf{Split} & \footnotesize \textbf{Start Date} & \footnotesize \textbf{End Date} & \footnotesize \textbf{Number of Data Points} & \footnotesize \textbf{Duration} \\
\hline
\footnotesize Training & \footnotesize 2020-07-02 16:35 & \footnotesize 2020-11-19 06:25 & \footnotesize 40,211 & \footnotesize 139 days, 13 hours, 50 minutes \\
\hline
\footnotesize Validation & \footnotesize 2020-11-19 09:35 & \footnotesize 2021-01-01 00:00 & \footnotesize 12,270 & \footnotesize 42 days, 14 hours, 25 minutes \\
\hline
\end{tabular}
\end{table}

\section{Simulations and Results}
This study employs an autoregressive approach for top-oil temperature estimation for the considered models. In this scenario, we assume that future top-oil temperature measurements are unavailable; therefore, the model's preceding estimations are used as inputs for subsequent predictions. The IEC 60076-7 standard’s differential equation for determining top-oil temperature is solved numerically using the finite difference method. By discretising the time interval into smaller sub-intervals $D_t$ and assuming the initial temperature $T_o$ and load $K$ constant within the interval, the finite difference method estimates the change in the oil temperature $DT_o$ at each time step:
\begin{equation*}
    {DT_o} = \frac{Dt}{k_{11}\tau_o} \left( \left[ \frac{1+K(t)^2\Psi}{1+\Psi} \right]^\chi  (\Delta{T_{or}}) - [T_o(t) - T_a(t)] \right).
\end{equation*}
At each time step, the top-oil temperature is determined by: 
\begin{equation}
    {T_o}_{(n)} = {T_o}_{(n-1)} + D{T_o}_{(n)},
\end{equation}
where ${T_o}_{(n-1)}$ is the previous top-oil temperature. The time step $D_t$ is chosen so that it is not less than half the smallest winding time constant $\tau_w$. For this investigation, numerous values of $D_t$ were considered within a sensible range.  Furthermore, because the information regarding the change of cooling stages was not available we assume that the power transformer does not go into forced air cooling mode but is constantly self-cooling through the natural convection of the oil. Hence when solving the equation we only consider the time constants related to ONAN cooling. 

Each of the deep learning-based models discussed in Section \ref{sec:methodology} has its own particular set of architecture-specific parameters. Initial experiments are conducted for each model to find the optimal learning rate, optimiser and batch size. Once the training parameters are identified and fixed, we employ a mixed approach of trial and error alongside a systematic grid search to determine the optimal configuration for each model. The explored hyperparameters for each model are presented in Table \ref{tab:hyperparameters}. Additionally, the models are trained using three different look-back windows of 2, 4, and 8 hours respectively. The look-back window represents the period of past samples that the model considers when making predictions.


\begin{table}[ht]
\renewcommand{\arraystretch}{1.3}
    \centering
    \caption{Hyperparameters and the tested values for each algorithm}
   \label{tab:hyperparameters}
   \begin{tabular}{|l|l|l|}
        \hline
        \textbf{Algorithm} & \textbf{Parameter} & \textbf{Values} \\ \hline
        \multirow{2}{*}{ANN} & number of neurons & [32, 64, 128]\\ \cline{2-3} & number of layers & [2, 4, 8] \\ \hline
        \multirow{2}{*}{TCN} & kernel size & [2, 4]\\ \cline{2-3} & number of filters & [8, 16, 32] \\ \hline
        \multirow{2}{*}{TiDE} & temporal decoder hidden & [8, 16]\\ \cline{2-3} & decoder output dimension &  [2, 4, 8] \\ \hline
    \end{tabular}
\end{table}

We approach the problem from both standard regression and quantile regression perspectives. In the typical regression context, the models are trained to optimize the mean absolute error. For quantile regression, we optimize the quantile loss. Overall, we consider estimating the $98\%$ conditional prediction interval bounded by the 0.01th and 0.99th conditional quantiles:
\begin{equation}
    {PI_{98}(\mathbf{x})} = [Q_{0.01}(\mathbf{x}), Q_{0.99}(\mathbf{x})].
\end{equation}
The models are compared using the mean absolute error and the mean squared error:
\begin{equation*}
MAE = \frac{1}{M}\sum_{t=1}^M{|y_t - \hat{y}_t|},
 \quad 
MSE = \frac{1}{M}\sum_{t=1}^M{(y_t - \hat{y}_t)^2},
\end{equation*}
where $M$ is the number of time steps in the predicted and actual top-oil temperature time series. Whilst the MAE is less sensitive to outliers, MSE allows us to infer if there are instances where models struggle with estimations.


In addition to estimating the top-oil temperature, we also investigate estimating both the top-oil temperature and the temperature rise over the ambient temperature $\Delta T$. The temperature rise is determined by the difference between the top-oil temperature and the corresponding ambient temperature at that specific time. Both predictions are made autoregressively, as previously elaborated. It is important to note that for the quantile regression models, the performance metrics are determined using the estimated 0.5th quantile. As discussed earlier, this represents the median tendency of the model’s estimations. Furthermore, when evaluating the prediction intervals it is important that we note two key metrics; the prediction interval width and the prediction interval coverage probability. The prediction interval width represents the mean width between the lower and upper estimated quantiles. This determines the uncertainty of the prediction intervals over the entire dataset, such that a narrower width represents confident estimations. The prediction interval coverage probability is the percentage of true values that fall within the estimated prediction interval. Ideally, a well-calibrated prediction interval has a narrow interval width and satisfies \eqref{prediction_interval_condition_satisfaction}. 

\subsection{Estimating top-oil temperature and the temperature rise over the ambient}
As discussed in the previous section, we first investigate the estimation of the top-oil temperature and the temperature rise over ambient. The models are trained to estimate these temperature values based on historical top-oil temperature, ambient temperature, and load measurements. During evaluation, the models generate one-time-step autoregressive estimations for the entire validation set. Table \ref{tab:model_training_parameters} presents the training parameters that were used for each deep learning model whilst Table \ref{tab:model_performance} reports the performance of the best configurations. We only report the top-oil temperature metrics for the IEC 60076-7 standard model since this model cannot be used directly to determine the temperature rise over ambient. For training the deep learning models, we used the Adam optimizer. In our experiments manually scaling the features to a comparable scale proved more effective than standardisation or robust scaling. 

\begin{table}[ht]
\centering
\renewcommand{\arraystretch}{1.5}
\caption{Training parameters for each model}
\label{tab:model_training_parameters}
\begin{tabular}{|l|l|l|l|}
    \hline
    \textbf{Model} & \textbf{Batch size} & \textbf{Max epochs} & \textbf{Learning rate}\\ \hline
    ANN & 256 & 4000 & 1e-5 \\ \hline
    TCN & 512 & 500 & 1e-4 \\ \hline
    TiDE & 512 & 100 & 1e-6 \\ \hline
\end{tabular}
\end{table}

 The best ANN performance is achieved with a model featuring 8 layers, 128 neurons per layer, and a look-back window of 4 hours. This configuration manages to achieve mean absolute errors of 1.49 $^{\circ}\text{C}$ for top-oil temperature and 1.53 $^{\circ}\text{C}$ for temperature rise over ambient. Deeper ANN configurations are shown to offer improved performance compared to shallower ones generally, however, the model's effectiveness could be further optimized by fine-tuning the interplay between network depth, width, and the input look-back window. 
 For our experiments, TCN models with a filter size (kernel) of 2 generally outperformed other configurations across various look-back window lengths and filter numbers. The best-performing configuration had a kernel size of 2, 16 filters, and a look-back window of 4 hours achieving mean absolute errors of 3.49 $^{\circ}\text{C}$ and 2.83 $^{\circ}\text{C}$ for the top-oil and temperature rise estimations, respectively. The optimal TiDE configuration featured an 8-dimensional output, temporal decoder layers of width 8, and a 4-hour look-back window. This configuration achieved mean absolute errors of 1.86 $^{\circ}\text{C}$ and 3.28 $^{\circ}\text{C}$ for the top-oil and temperature rise estimations, respectively. All the proposed models show a better performance than the standard model over the validation set. Overall, the ANN demonstrates the best performance for both top-oil and temperature rise estimations, with the lowest values for both MAE and MSE. It is important to consider that the systematic grid search for the ANN compared to the more complex architectures because they take less time to train. Also, the TCN and TiDE are typically used for long-term time-series forecasting, estimating many timesteps into the future with a single prediction. Ultimately, the grid search for each of these models was not exhaustive, additional exploration parameters may reveal even better configurations, balancing model complexity with generalization ability for enhanced predictive accuracy. 

\begin{table}[ht]
\renewcommand{\arraystretch}{1.5}
\centering
\caption{Perfomance of models}
\label{tab:model_performance}
\begin{tabular}{|l|l|l|l|l|}
    \hline
  \multirow{2}{*}{\textbf{Model}} & \multicolumn{2}{l|}{\textbf{Top-oil}} &  \multicolumn{2}{l|}{\textbf{Temp. rise}} \\ \cline{2-5}
    &  MAE & MSE & MAE & MSE \\ \hline
    IEC 60076-7 & 5.51 & 57.25 & - & - \\ \hline
    ANN & 1.49 & 3.49 & 1.53 & 3.70 \\ \hline
    TCN & 3.49 & 18.18 & 2.83 & 14.19 \\ \hline
    TiDE & 1.86 & 6.02 & 3.28 & 18.47 \\ \hline
\end{tabular}
\end{table}

\begin{figure}[ht]
\centering
\includegraphics[width=0.9\textwidth]{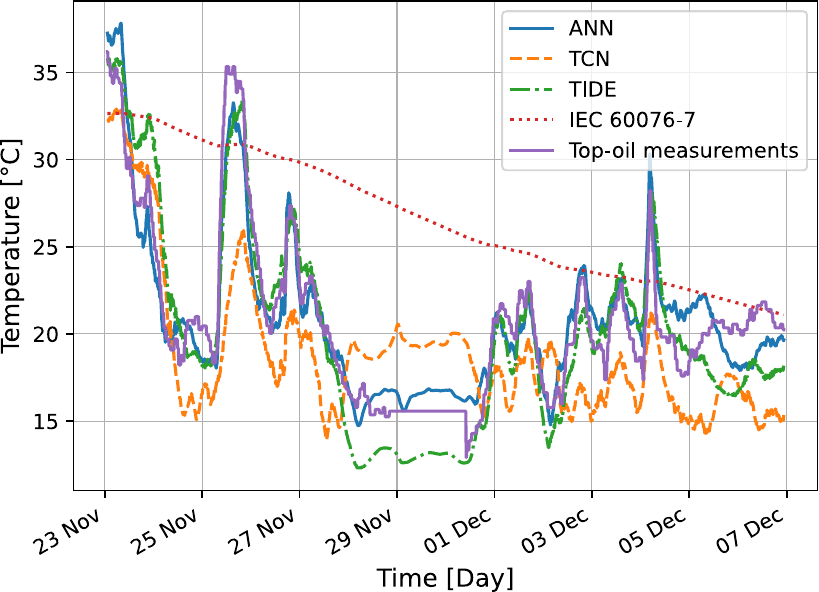}
\caption{Top-oil temperature estimates from the proposed models against measurements.}
\label{fig_2}
\end{figure}

Figure~\ref{fig_2} compares top-oil temperature estimations by each of the models considered in this study against the top-oil measurements in the validation dataset. In comparison to the standard model the neural network-based models all closely follow the temperature measurements, each with a varying degree of deviation. It can be observed that for each model there are instances of under-estimation and over-estimation of the oil temperature. In some cases, the model can capture the general trend of the temperature change but the amplitude of the temperature change is not fully captured. These under-estimations and over-estimations are of particular interest to the application of power transformer condition monitoring since it is a safety-critical application. Failure to anticipate higher top-oil temperatures can lead to catastrophic failure and damage to equipment. Conversely, incorrectly anticipating higher oil temperatures can also lead to under-utilization of equipment in cases where demand is high. Therefore, there is a need for a mechanism that can provide a range of possible temperature values rather than a point estimation. 

\subsection{Estimating top-oil temperature with prediction intervals}
Prediction intervals provide a range of values for top-oil temperature estimations through predicted conditional quantiles, offering insight into estimation uncertainty. Table \ref{tab:model_performance_quantile_regression} presents the performance of the best configurations for each model type and Table \ref{tab:model_training_parameters_quantile_regression} shows the corresponding training parameters. Similar to the previous models we employed the Adam optimiser and manual scaling for features. The ANN exhibits the best performance, achieving a mean absolute error of 1.85 $^{\circ}\text{C}$ but with much lower coverage compared to the other models. Conversely, the best TCN configuration has the lowest performance of 2.86 $^{\circ}\text{C}$, but has the highest mean coverage of 0.77. The TiDE model has comparable performance to the other models, with a top-oil mean absolute error of 2.20 $^{\circ}\text{C}$ and a sufficiently adequate mean coverage of 0.73. Whilst an ANN configuration achieved the lowest error, the limited coverage is insufficient for our application. As noted earlier, an ideal well-calibrated model should have a high coverage and narrow mean prediction interval width. This is quite difficult to achieve as the model is not directly trained to optimize these properties.

\begin{table}[ht]
\centering
\renewcommand{\arraystretch}{1.5}
\caption{Training parameters for each model}
\label{tab:model_training_parameters_quantile_regression}
\begin{tabular}{|l|l|l|l|}
    \hline
    \textbf{Model} & \textbf{Batch size} & \textbf{Max epochs} & \textbf{Learning rate}\\
    \hline
    ANN & 256 & 4000 & 1e-5 \\ \hline
    TCN & 512 & 500 & 1e-4 \\ \hline
    TiDE & 512 & 100 & 1e-5 \\ \hline
\end{tabular}
\end{table}

\begin{table}[ht]
\renewcommand{\arraystretch}{1.5}
\centering
\caption{Performance of models trained on quantile loss}
\label{tab:model_performance_quantile_regression}
\begin{tabular}{|l|m{2cm}|m{2cm}|m{2cm}|m{3cm}|}
    \hline
    \textbf{Model} & \textbf{Top-oil MAE} & \textbf{Top-oil MSE} & \textbf{Mean coverage} & \textbf{Mean interval width} \\
    \hline
    ANN & 1.85 & 5.00 & 0.16 & 1.11 \\ \hline
    TCN & 2.86 & 13.42 & 0.77 & 9.71 \\ \hline
    TiDE & 2.20 & 8.71 & 0.73 & 6.10 \\ \hline
\end{tabular}
\end{table}

\begin{figure}[ht]
\centering
\includegraphics[width=0.9\textwidth]{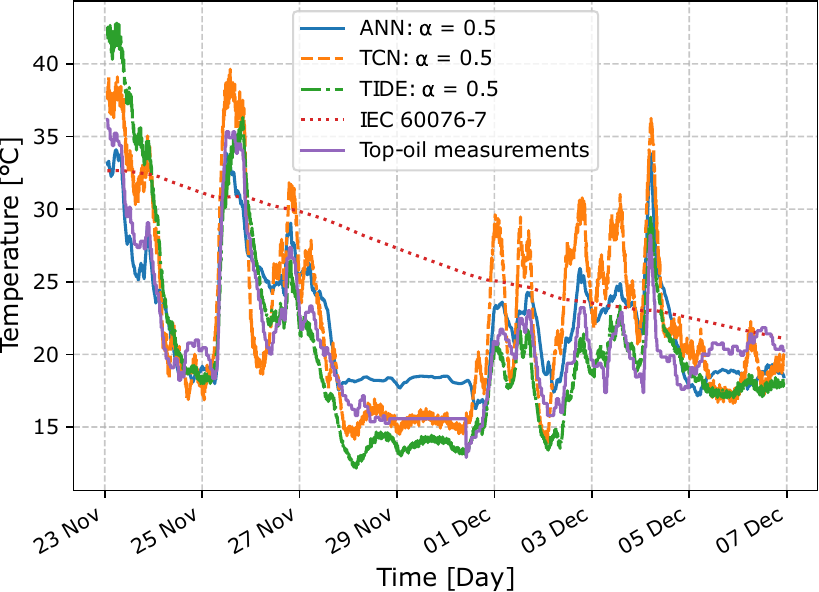}
\caption{Top-oil temperature estimates for each model at the conditional quantile  $\alpha = 0.5$ against measurements.}
\label{model_estimations_comparisons_quantile_models}
\end{figure}

Figure \ref{model_estimations_comparisons_quantile_models} compares the estimations for each model at the conditional quantile  $\alpha = 0.5$ against measurements. For this snapshot, the neural network models closely match the temperature measurements with varying degrees of error, while the IEC model exhibits the largest deviations. The TCN experiences large deviations at the peaks whilst the TiDE model underestimates the temperature in some instances. 

\begin{figure}[ht]
\centering
\includegraphics[width=0.9\textwidth]{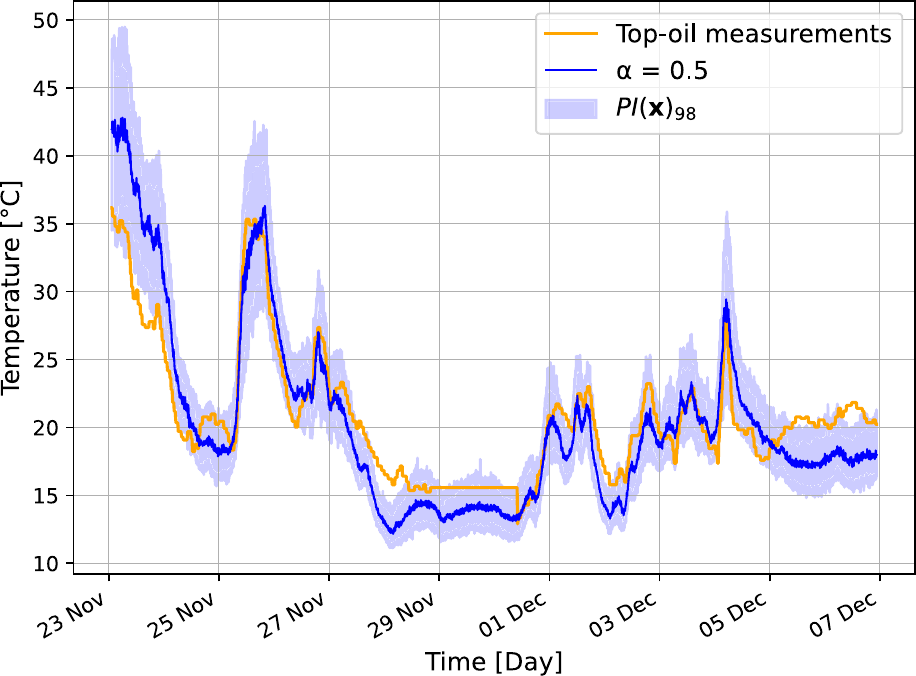}
\caption{ Top-oil temperature measurements against the estimated conditional quantile at $ \alpha$ = 0.5 and the interval $PI_{98}(\mathbf{x})$.}
\label{fig_3}
\end{figure}

Figure \ref{fig_3} shows the top-oil temperature measurements against the TiDE model's estimated conditional quantile at $\alpha = 0.5$ and the conditional prediction interval $PI_{98}(\mathbf{x})$ bounded by the 0.01th and 0.99th conditional quantiles. Instead of having a point estimation, we can determine the minimum and maximum expected oil temperature at a given point in time. The shaded area represents the range of expected values with a large distance between the conditions quantiles representing a large uncertainty in the prediction. From this example, we can determine that the temperature measurements fairly fall into the prediction interval. As presented in Table \ref{tab:model_performance_quantile_regression}, approximately $73\%$ of the top-oil temperature measurements in the validation set fall into the conditional prediction interval. However, the estimated conditional prediction interval fails to satisfy~\eqref{prediction_interval_condition_satisfaction}. As previously discussed, quantile loss effectively estimates quantiles, but it does not explicitly optimize for prediction interval coverage. To address this limitation, researchers have combined quantile regression with other methods capable of providing coverage guarantees, such as conformalised prediction intervals  \cite{romano2019conformalized}. 

\section{Discussion and Conclusion}
Monitoring the temperature distribution inside power transformers is essential to track their lifetimes. The deterioration of the oil-paper insulation system is one of the indicators of the remaining life of power transformers. As the insulation system is subjected to electrical currents and temperature fluctuations, the temperature changes need to be properly monitored. Conventional methods, like the IEEE and IEC 60076-7 standards, estimate the top-oil and hot-spot temperatures that can be considered indicators for the transformer's remaining lifetime. However, these models are often inaccurate and depend specifically on the power transformer properties. In our work, we propose several machine learning methods for time series forecasting to estimate the top-oil temperature. We adopt the  Time-series Dense Encoder (TiDE) and the Temporal Convolutional Networks (TCN) architectures from the Darts library, which allows us to investigate the applicability of advanced time-series forecasting architectures for our problem along with traditional artificial neural networks (ANNs). The proposed models are shown to outperform the IEC model with the best model being the ANN achieving an error of 1.49 $^{\circ}\text{C}$. Moreover, we investigate how quantile regression can be used to determine the uncertainty of temperature estimations. Conditional quantile pairs can be used to construct a prediction interval to which we expect the top-oil temperature. Instead of having a point estimation for each forecast, we have maximum and minimum possible values for the expected top-oil temperature. The best-performing model was able to estimate conditions quantiles that cover an acceptable number of samples in the validation data. The results achieved are already satisfactory, and more improvements can be achieved in this direction. The following steps would include a comparison with the novel models in machine learning that combine data with physics, like physics-informed neural networks. These networks can exploit the physics expressed in the equation of the interested system and therefore can generalize to unseen cases in training data. As previous work was carried out in this area of predicting the temperature distribution inside power transformers \cite{ref5, fede1, fede2, fede3}, we would like to further extend the comparison of predictions of purely data-driven machine learning methods to PINNs. 

\section*{Acknowledgment}

This work is supported by the Vinnova Program for Advanced and Innovative Digitalisation (Ref. Num. 2023-00241) and Vinnova Program for Circular and Biobased Economy (Ref. Num. 2021-03748) and partially supported by the Wallenberg AI, Autonomous Systems and Software Program (WASP) funded by the Knut and Alice Wallenberg Foundation.

\bibliographystyle{elsarticle-num} 
\bibliography{references}



\end{document}